\documentclass{svproc}
\usepackage{url}
\usepackage{graphicx}
\usepackage[namelimits]{amsmath}
\usepackage{multirow}
\usepackage{subfigure}

\begin{document}
\mainmatter              
\title{Video Face Recognition System: RetinaFace-mnet-faster and Secondary Search}

\titlerunning{Video Face Recognition System}  
%
%
%
%
\author{Qian Li\inst{1,2} \and Nan Guo\inst{1} \and Xiaochun Ye\inst{1,2} \and Dongrui Fan\inst{1,2} \and Zhimin Tang\inst{1,2}
}
\authorrunning{Qian Li et al.}
\tocauthor{Qian Li, Nan Guo, Xiaochun Ye, Dongrui Fan, and Zhimin Tang}
%
%
\institute{State Key Laboratory of Computer Architecture, Institute of Computing Technology,\\
Chinese Academy of Sciences, Beijing, China\\
\email{\{liqian18s,guonan,yexiaochun,fandr,tang\}@ict.ac.cn}\\
\and
University of Chinese Academy of Sciences, Beijing, China\\
\email{\{liqian18s,yexiaochun,fandr,tang\}@ict.ac.cn}}

\maketitle              

\begin{abstract}
Face recognition is widely used in the scene. However, different visual environments require different methods, and face recognition has a difficulty in complex environments. Therefore, this paper mainly experiments complex faces in the video. First, we design an image pre-processing module for fuzzy scene or under-exposed faces to enhance images. Our experimental results demonstrate that effective images pre-processing improves the accuracy of 0.11\%, 0.2\% and 1.4\% on LFW, WIDER FACE and our datasets, respectively. Second, we propose RetinacFace-mnet-faster for detection and a confidence threshold specification for face recognition, reducing the lost rate. Our experimental results show that our RetinaFace-mnet-faster for 640$\times$480 resolution on the Tesla P40 and single-thread improve speed of 16.7\% and 70.2\%, respectively. Finally, we design secondary search mechanism with HNSW to improve performance. Ours is suitable for large-scale datasets, and experimental results show that our method is 82\% faster than the violent retrieval for the single-frame detection.
\end{abstract}
\section{Introduction}
With the development of face recognition technology, researchers propose many methods \cite{TURK1991Eigenfaces} \cite{Paysan2009A} \cite{Valentin2004Connectionist} \cite{6909616} \cite{Yi2014Deep}. \cite{TURK1991Eigenfaces} based on traditional face features recognizes faces. Based on matching, \cite{Paysan2009A} completes similarity measurement, dynamically. \cite{Valentin2004Connectionist} proposes a 5-dimensions retrieval method, in which datasets are from 50 components, performing better than simple images. \cite{6909616} based on deep learning exploits Soft-Max classification. However, \cite{TURK1991Eigenfaces} \cite{Valentin2004Connectionist} perform poor in face recognition, \cite{6909616} \cite{Yi2014Deep} \cite{Yi2014Learning} \cite{Chopra2005Learning} \cite{Fan2014Learning} cost more time. Therefore, we improve the speed and accuracy of face recognition by optimizing the face detection.

Face recognition mainly includes face detection, localization, face pre-processing and recognition. Many face images methods \cite{Choi2011A} \cite{Dharavath2014Improving} \cite{Zhang2011Image} \cite{Jridi2017Rapid} mainly focus on diverse environments around faces, different lighting and cameras, ensuring the sub-sequent face recognition more robust. 

Compared with static faces, face recognition methods \cite{Guo2005Face} \cite{Gorodnichy2005Video} \cite{Park20073D} \cite{Park20053D} in the video are more difficult because of poor quality, small scales and complex scenes. The quality of frames is poor, the face is small and blurry. The face scene is insufficient,  resulting in the difficulty in recognizing. In this work, we mainly focus on pre-processing face images, face detection network and multiple searches to improve performance. Therefore, our attributions as follow:
\begin{quote}
\begin{itemize}
\item We design pre-processing module in complex scenes. According to the median filter, we enhance the quality of faces, improving efficiency under luminous conditions. After removing more noise, we restore the image by the Wiener filter to the greatest extent. In this work,  we improve performance of 0.11\% on the LFW.
\item According to stride and threshold, we propose RetinaFace-mnet-faster. Our experimental results show that we improve the speed 1.7 relative to the original network on Tesla P40. The acceleration ratio is 1.2 on a single-threaded CPU. The recognition AP achieves 78.4\%, increasing by 0.2\% on WIDERFACE.
\item We design a secondary retrieval method on HNSW library for large-scale features, quickly searching and matching. Experiments show that the detection speed of per-frame is 82\% higher than the brute force method on videos.
\end{itemize}
\end{quote}

\section{Related Works}
\textbf{Face Recognition.} Researchers have proposed many face recognition algorithms \cite{TURK1991Eigenfaces} \cite{Paysan2009A} \cite{Valentin2004Connectionist} \cite{6909616} for different problems \cite{cootes2001active} \cite{Doll2010Cascaded} \cite{Yi2013Deep} \cite{Xu20172}. Rather than traditional face recognition algorithms, \cite{cootes2001active} \cite{Xu2004Automatic} \cite{Luo2018A} based on the geometric features and \cite{Yue2017Computationally} \cite{Bodla2017Deep} based on the template-based methods, we combine traditional methods and deep learning methods to improve face recognition performance. Face alignment plays an important role in face recognition, adjusting the face and key points by angle better and normalized faces to obtain the more concentrated features. According to similarity, the bigger the distance is, the smaller the similarity is. Therefore, we design the secondary search to improve speed and accuracy.

\textbf{Face Recognition in Videos.} Based on face recognition of static images, face recognition on videos obtains face name and similarity. Researchers propose many algorithms \cite{Guo2005Face} \cite{Gorodnichy2005Video} \cite{Park20073D} \cite{Park20053D}. Compared with static images, frames in the video have much more challenges, such as complex scenes, tiny scales, and blurring, etc. Because of the relevance between adjacent frames, \cite{Wheeler2007Multi} \cite{Dhamecha2016On} \cite{Selvaganesan2016Unsupervised} further increase accuracy of face recognition on videos. However, the real-time performance poorly.

\textbf{Image Enhancement.} The pre-processing images \cite{Wahl1988Image} \cite{Fossum1987Charge} \cite{Bieniecki2007Image} \cite{Dharavath2014Improving} are different according to different environment, enhancing the visual features and obtaining more key information. \cite{Pandey2017Enhancing} \cite{Yelmanova2016Automatic} \cite{Li2018Image} enhance contrast, \cite{Moujahid2018Feature} \cite{Jia2009Face} \cite{Dube2014Does} use spatial filter to enhance features, and \cite{Srinivas2011Sparsity} exploits density slices to convert gray. As we known, compressing high dynamic range scenes is a possible solution for current devices with the limited dynamic ranges (LDR). \cite{Provenzi2009Perceptual} \cite{Chude2013Illumination} \cite{Rahman2011Performance} propose global histogram correction methods, gamma adjustment, log compression, histogram equalization etc. While these methods may lose or unenhanced some important features. Therefore, we use a wavelet-based contrast enhancement module to improve the performance of face detection. We mainly adjust the contrast, hue and saturation and use wavelet-based threshold to enhance the image.

\section{Our Approach}
We mainly optimize face detection and face recognition networks, and this paper focuses on face image enhancement in diverse scenes to improve the accuracy, and optimizes the RetinaFace-mnet from batch sizes, scales, strides, etc. to improve the speed of face detection. Finally, according to the secondary search, we improve the accuracy of face recognition. The following mainly details image pre-processing and RetinaFace-mnet-faster.
\begin{figure*}[t]
\centering
\includegraphics[width=1.0\textwidth]{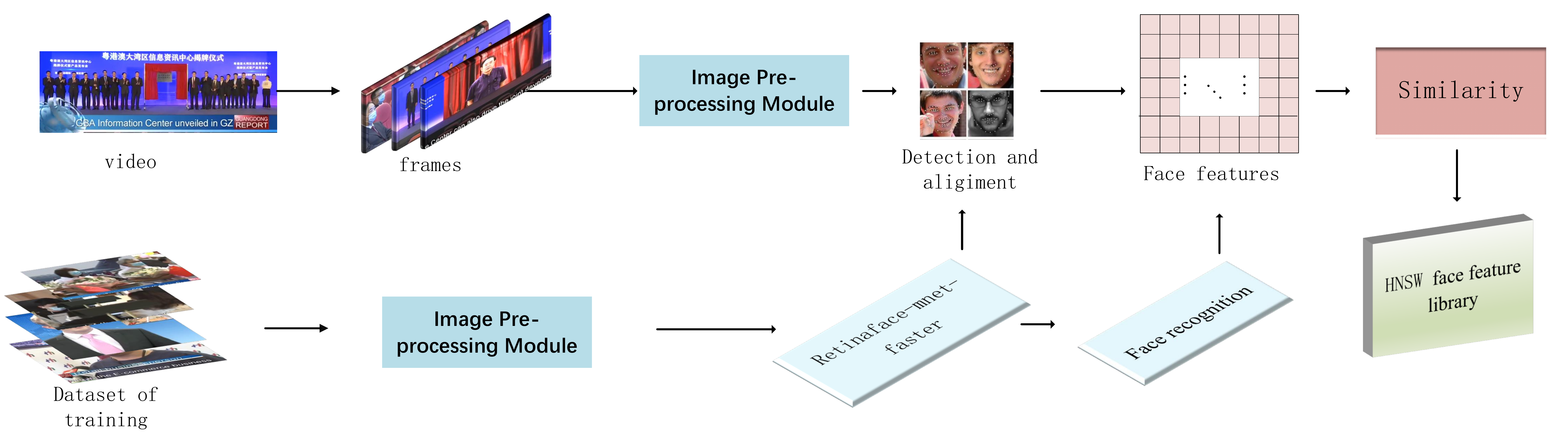} 
\caption{Our video face recognition system. It mainly contains two paths. The first path uses the training set to get model. After pre-processing image, the Retinaface-mnet-faster trains face images. Then, we cut faces obtained by Retinaface-mnet-faster, and recognition model trains faces, and two main models (detection and recognition models) are obtained. Another path recognizes faces on videos. The extracted frames are pre-processed by image pre-processing module, detecting by face detection model, and alignment. Then, we use the face recognition model to extract face features. Finally, we compute the similarity to complete face recognition between the extracted features and features of the face feature library. The face feature library is constructed in advance.}
\label{figure1}
\end{figure*}

\subsection{Image Pre-processing Module} 
\begin{figure*}[t]
\centering
\includegraphics[width=1.0\textwidth]{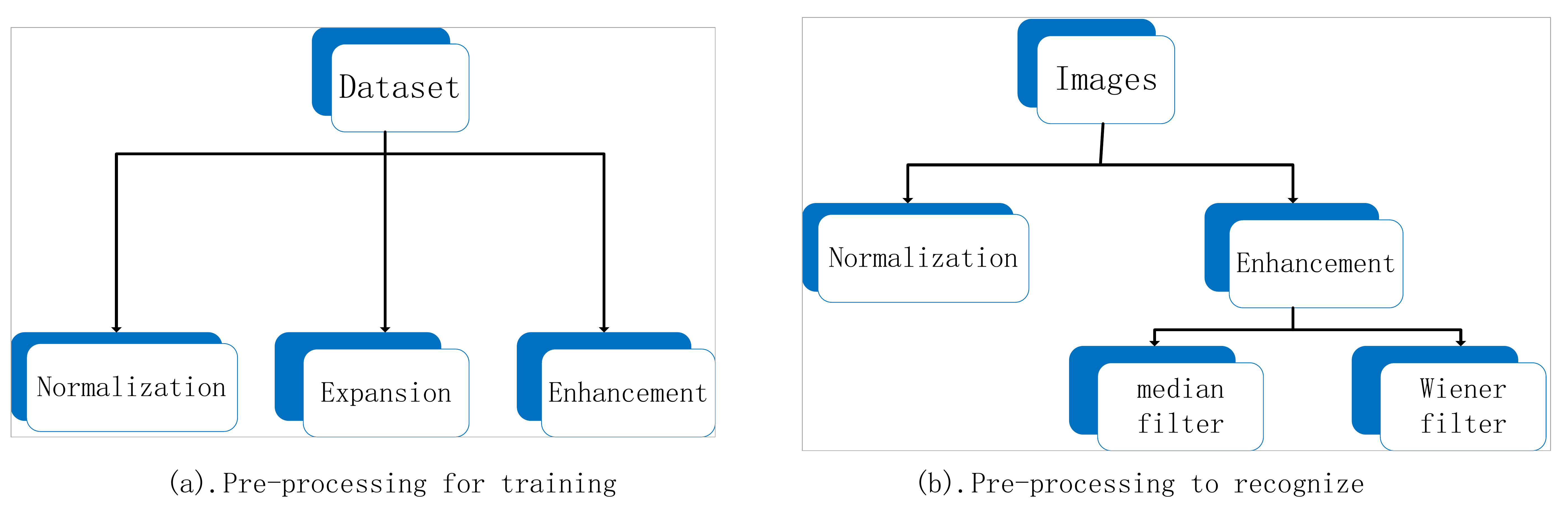}
\caption{Our image pre-processing modules for training and recognizing.}
\label{figure2}
\end{figure*}

As shown in Figure \ref{figure2} (a), our pre-processing on face images mainly focuses on diverse scenes for face recognition tasks in the video. The brightness is different because of diverse environment. Some faces are low-quality and blurred. This paper mainly pre-processes images to improve performance of face recognition. Therefore, we design this module for training/test datasets. The pre-processing between the training and test sets have the same and different parts because of different emphasis. We focus on the number, enhancement and normalization for the training set. As shown in Figure \ref{figure2} (b), for test dataset, we mainly focus on blurry, insufficient lighting and excessive noise, eliminating the insufficiency in the image. According to strategies, we improve the accuracy in the whole video face recognition system.

\textbf{Scale Setting.} For the training set, we set multi-scales transformation strategies. The size of input is 320$\times$320. According to the scale transformation parameters $scales$, $scales$ are set to $[0.2,0.4,0.6,0.8,1]$, a random transformation is performed on each input image, such as randomly performing $320 /scales$, we can obtain the multi-scales images, $[320,400,533,800,1600]$. Then, we transform the position of faces and landmarks, obtaining a complete scale-transformed images. We crop 25 times in every image as 320$\times$320, and transform coordinates of position and landmarks for cropped images. Finally, we judge the cropped images whether the image is face. Afterwards, the cropped images are enhanced, normalized etc., obtaining the final input images, position coordinates and key point coordinates of the face frame.

\textbf{Image Enhancement.} For face recognition, we use the median filter to eliminate the noise. The median filter has a good processing effect on the night pictures. Then, we exploit the Wiener filter to restore the image to the greatest extent, improving the recognition effect of the blurred face, and we normalize the processed image.

\subsection{RetinaFace-mnet-faster}
RetinaFace-mnet-faster mainly optimizes the face detection (RetinaFace-mnet) to improve the accuracy, speed, recall rate, and decrease number of false detections. According to \cite{Lin2017Focal}, \cite{Deng2019face2} proposes the RetinaFace-mnet to improve the performance on multi-scale objects. The RetinaFace-mnet based on light-weight MobileNet is faster and more light-weight. RetinaFace-mnet reduces false detection by SSH module which includes a context module and a detection module with a convolution layer.

RetinaFace-mnet contains three branches, corresponding to the three-layer feature pyramids, for which three strides are set as 8,16, and 32, respectively, improving the performance for multi-scale objects (small, medium and large). Then, the author sets two scales coefficients in each stride, so that every feature can obtain two anchor boxes. In our work, we mainly optimize three aspects. First, we improve the speed by changing the input image resolution and the batch sizes. Then, based on the appropriate resolution, we adjust the branch with different strides, and remove the key point detection of stride8 and stride32 and stride16. Finally, we adjust the threshold according to the recall rate to reduce the missed detection rate.
\subsection{ HNSW secondary search feature}
\begin{figure*}[t]
\centering
\includegraphics[width=1.0\textwidth]{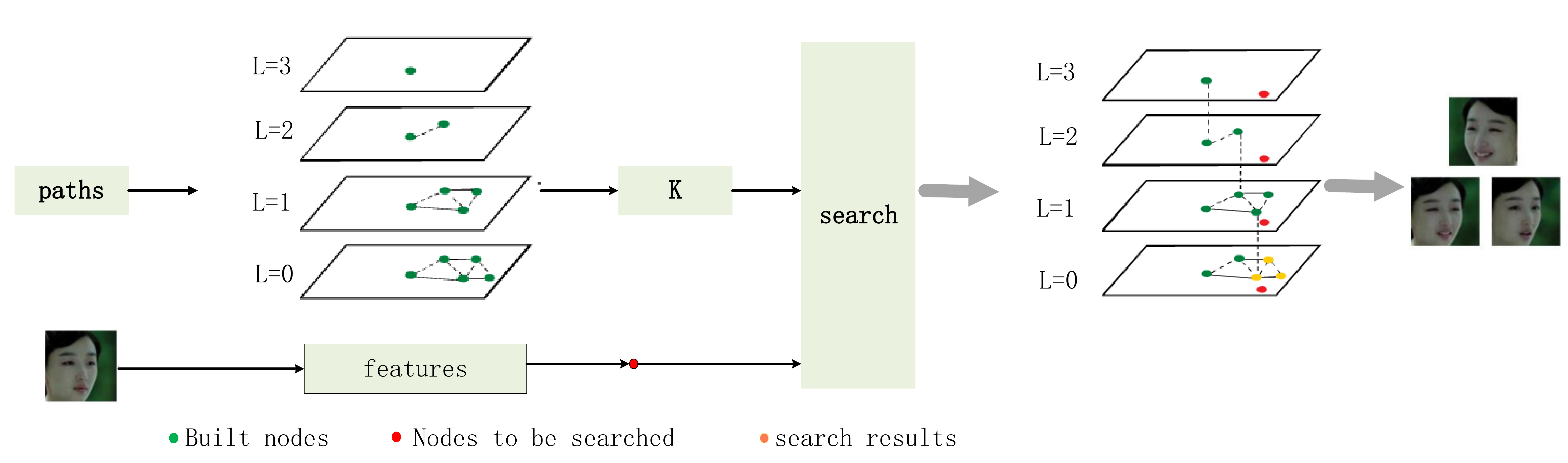}
\caption{Similar feature retrieval of HNSW. The search node $q$ is represented in red (may be a node in the non-graph), the established nodes are represented in green, and the search results are represented in yellow. The parameters include the query point $q$, the starting node $ep$, and the number of nearest neighbors to be found $ef$. The result obtains $ef$ nearest neighbors of node $q$ in this layer}
\label{figure3}
\end{figure*}

NSW alleviates the problem of approximate $k$ nearest neighbor search in the metric space. However, due to the characteristics of divergent search, the time complexity is very high. Based on the idea of jumping the table, \cite{Malkov2016Efficient} proposes Hierarchical-NSW (HNSW) to decrease the time. As shown in Figure \ref{figure3}, HNSW uses a multi-layers graph. The higher-level graph contains fewer vertices, the lower the degree of each vertex is. And the lower-level graphs contain more vertices, the degree of each vertex is also greater. The algorithm can implement balanced distribution because of the similarity of skip table.
\begin{figure*}[t]
\centering
\includegraphics[width=1.0\textwidth]{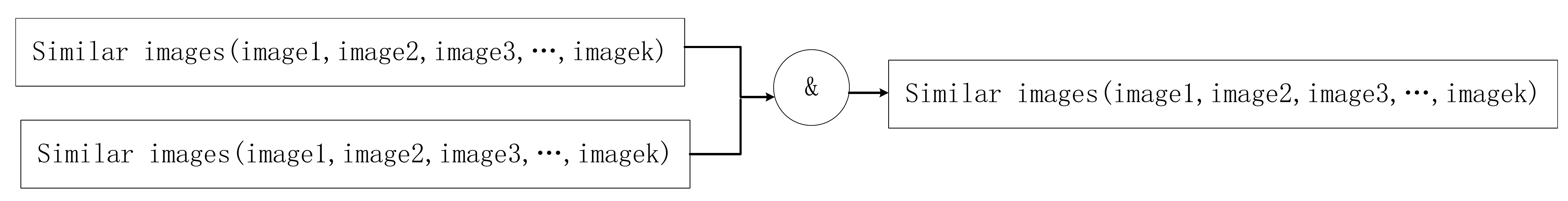}
\caption{Schematic diagram of face search method based on quadratic search.}
\label{figure4}
\end{figure*}
\section{Experiments and Results}
 This experiment mainly evaluates the face recognition algorithm from the recall rate, false detection rate and speed.
 
\textbf{HNSW retrieval.}  As shown in Figure \ref{figure3}, the retrieval processing is detailed. As shown in Figure \ref{figure4}, to improve the recall rate and precision rate of the search, we propose the secondary search method. After finding the $K$ similar pictures for the first time, the same method is used to perform the first search. According to $logical$ operation between the first results and the secondary results, we obtain higher precision and recall. Logical operation can obtain higher precision for different application scenes.

\textbf{DataSets.} We mainly experiment WIDER FACE, LFW and our datasets. WIDER FACE has 32,203 images with 393,703 faces of 61 events with different angles and poses. For each type of event, we randomly select 40\%/10\%/50\% as training/validation/test. The evaluation is same with PASCAL VOC in WIDERFACE. We evaluate the RetinaFace-mnet-faster on WIDER FACE. LFW (labeled Faces in the Wild) is widely used face dataset in the unrestricted scenes. The LFW contains 13233 faces and 5749 persons. Each face is marked with the name of the person. 1680 persons appear in two or more different photos. Our datasets with 3000 persons of 15000 faces are from practical applications.
\subsection{Retinafacce-mnet-faster}
\begin{figure*}[t]
\centering
\includegraphics[width=0.9\textwidth]{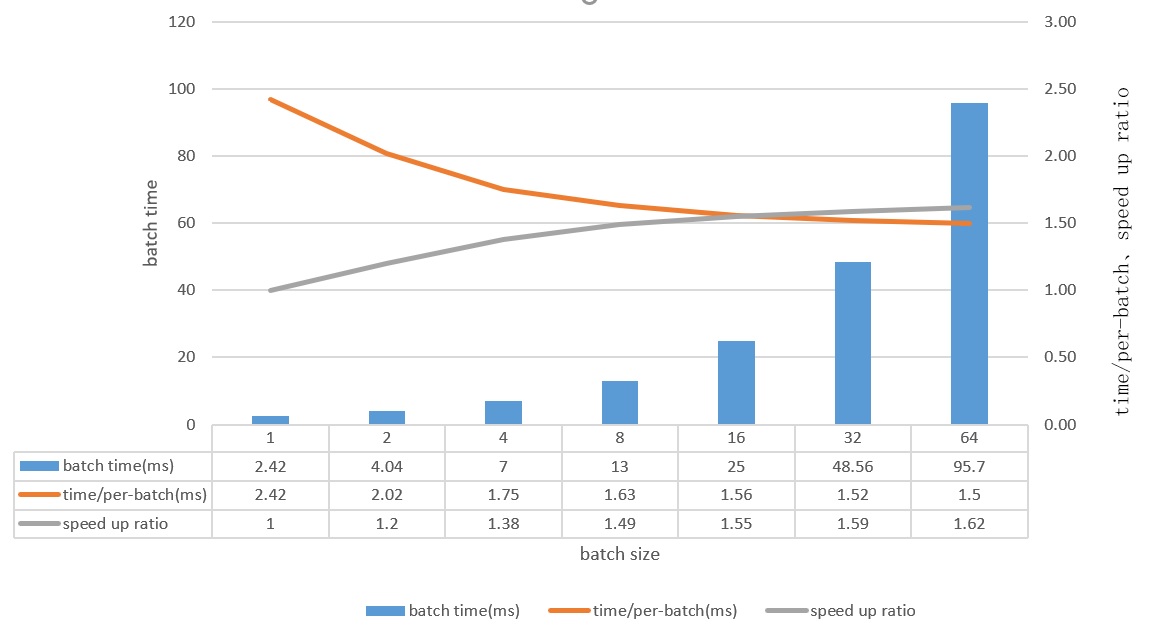}
\caption{Inference time performance evaluation of TensorRT 5.1.5 Neural network.}
\label{speedup}
\end{figure*}
The speed of Retinafacce-mnet is evaluated through the TensorRT acceleration platform. As shown in Figure \ref{speedup}, if batch calculation is performed, the speed-up ratio increases with the increase of the batch, but the effect is not obvious on TensorRT 5.1.5. We find different resolutions have a great impact because of batch sizes. 

Our input resolution is 320$\times$320. RetinaFace-mnet has 3 detection branches on the feature pyramid, corresponding to different strides: 32, 16, and 8, respectively. These three detection branches are different receptive fields. As the resolution of input image becomes smaller, we adjust the strides of the three detection branches. Due to the particularity of face images in monitoring and other scenes, network has higher requirements for small-scale faces. We remove the detection branch with stride of 32, and select the right lower sampling stride as 4 and 8, corresponding to the size of the feature pyramid 80$\times$80 and 40$\times$40, respectively. Since the key point detection of small faces has little effect on the alignment of faces on downstream tasks, we use the regression task of frugal key point detection to improve the detection speed, decreasing the detection time for a specific small face.

\textbf{Training Setting.} As the threshold decreases, the network is more likely to detect faces, while there are many false faces. Therefore, the appropriate threshold has a great impact on face detection of different situations. At the same time, the NMS threshold is related to overlapped face areas. When training, we set the NMS-threshold of positive and negative examples to greater than 0.5 and less than 0.3, respectively. We use the standard OHEM to alleviate the imbalance between positive and negative examples, achieving the ratio of 1:3. We use the SGD optimizer to train RetinaFace-nmet-faster, the momentum, weight decay, batch size and the initial learning rate are set as 0.9, 0.0005, 8$\times$4, and 0.01, and the learning rate becomes 0.001 after 5 epochs.
\subsection{ Evaluations}
\textbf{The importance of RetinaFace-mnet-faster.} As detailed in Table \ref{table1}, the speed of RetinaFace-mnet-faster is faster than the RetinaFace-mnet, especifically for low resolution. the speed of our RetinaFace-mnet-faster on the Tesla P40 is 1.7, which is 16.7\% higher than the RetinaFace-mnet  for the 640$\times$480. The acceleration ratio on the single-thread is 1.2, and the speed is increased by 70\%. We think that according to the appropriate threshold and the strides varying with the resolution, we significantly improve the performance.
\begin{table}
\small
\caption{Comparison RetinaFace-mnet with RetinaFace-mnet-faster for speed of detection.}
\label{table1}
\resizebox{0.95\columnwidth}{!}{ 
\smallskip\begin{tabular}{l|l|l|l}
\hline
Backbones&
VGA(640$\times$480)&
HD(1920$\times$1080)&
4K(4096$\times$2160)
\\
\hline
RetinaFace-mnet(GPU-TeslaP40)&1.4&6.1&25.6\\

RetinaFace-mnet(CPU-1)&17.2&130.4	&-\\

RetinaFace-mnet-faster(CPU-1)&10.1&92.3&-\\

RetinaFace-mnet-faster(GPU-TeslaP40)&1.2&4.6&22.1\\
\hline
\end{tabular}
}
\end{table}

\begin{figure*}[t]
\centering
\includegraphics[width=1.0\textwidth]{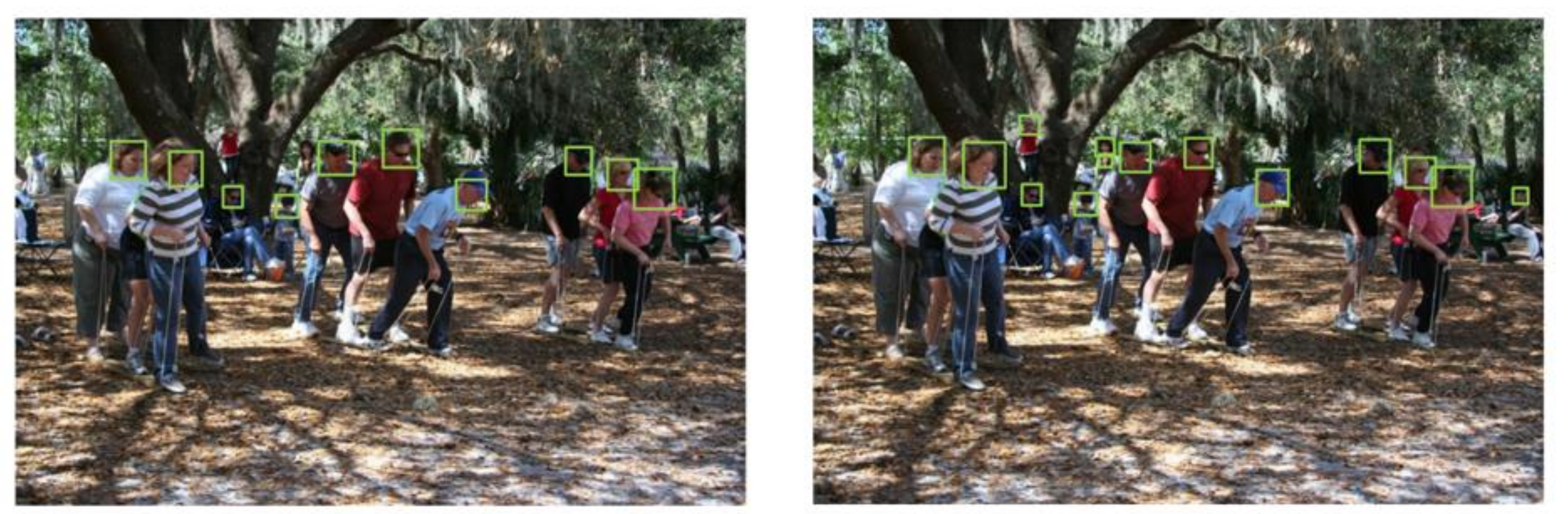}
\caption{Comparison Retinaface-mnet with Retinaface-mnet-faster. The figure in the left is the effect of the Retinaface-mnet, and the figure in the right is the effect of the Retinaface-mnet-faster. We find the performance of the Retinaface-mnet-faster is better than Retinaface-mnet.}
\label{figure4}
\end{figure*}

A shown in Table \ref{table2}, the AP of RetinaFace-mnet-faster on the WIDER FACE-Hard Set achieves 78.4\%, which is 0.2\% higher than RetinaFace-mnet.
\begin{table}
\small
\centering
\caption{Comparison Retinaface-mnet with Retinaface-mnet-faster on the WIDER FACE for accuracy.}
\label{table2}
\resizebox{0.5\columnwidth}{!}{ 
\smallskip\begin{tabular}{l|l}
\hline
Backbones&
WIDER FACE-Hard Set(Ap)
\\
\hline
RetinaFace-mnet&78.2\%\\

RetinaFace-mnet-faster&78.4\%\\
\hline
\end{tabular}
}
\end{table}

\textbf{The Importance of Secondary Search.} We mainly evaluate HNSW on precision and speed. We experiment unrestricted scene videos in surveillance and driving recorders. While building library, the time is 53.8s. As shown in Table \ref{table3}, the detection time of the per-frame is 12.2504 ms for 154 simple videos (the average number of faces in a single frame is less than 5). The one is 31.1067ms for the random 300 complex videos (the average number of faces in a single frame is greater than 5). Compared with the brute force search method, ours improves the speed greatly.
\begin{table}
\small
\centering
\caption{Evaluation on the retrieval speed of HNSW face features.}
\label{table3}
\begin{tabular}{l|l|l|l|l}
\hline
Test videos&\multicolumn{2}{c|}{Simple Videos}&\multicolumn{2}{c}{Complex Videos}\\
\hline
Testset&\multicolumn{2}{c|}{154 videos} &\multicolumn{2}{c}{300 videos}\\
\hline
\multirow{2}{*}{per-frame(ms)}&HNSW&12.2504&HNSW&31.1067\\
&Violence&22.3012&Violence&36.3892\\
\hline
\end{tabular}
\end{table}
\subsection{ Experimental analysis}
\textbf{Based on LFW dataset.} We test a pair of photos whether they are the same person. We evaluate the performance of face recognition by selected randomly 6000 pairs. As shown in Figure \ref{table46} (a), the method combing RetinaFace-mnet with ArcFace performs well on LFW, and the method combing RetinaFace-mnet-faster with ArcFace performs 0.11\% better than the one. Our RetinaFace-mnet-faster improves performance.

\begin{figure}[htbp] 
 \scriptsize

\subfigure[Results on LFW dataset.]
{
  \begin{minipage}[c]{0.5\textwidth} 
  \centering
    \begin{tabular}{l|l}
    \hline
			Methods&LFW Accuracy\\
			\hline
			RetinaFace-mnet+ArcFace&99.50\\
			RetinaFace-mnet-faster+ArcFace&99.61\\
			\hline
    \end{tabular}    
  \end{minipage}  
  }
\subfigure[Results on our dataset.]
{
  \begin{minipage}[c]{0.5\textwidth} 
  \centering
   \begin{tabular}{l|l}
\hline
Methods&Accuracy(\%)\\
\hline
RetinaFace-mnet+ArcFace&58.7\\
RetinaFace-mnet-faster+ArcFace&60.1\\
\hline
\end{tabular}  
  \end{minipage}  
  }
  
 \label{table46}
  \caption{Comparison of different datasets for RetinaFace-mnet+ArcFace and RetinaFace-mnet-faster+ArcFace. We find that our RetinaFace-mnet-faster detects better.}
\end{figure}

\begin{table}
\small
\begin{center}
\caption{Results on LFW dataset for different thresholds.}
\label{table5}
\resizebox{0.7\columnwidth}{!}{ 
\smallskip\begin{tabular}{l|l|l|l|l|l|l|l}
\hline
threshold(t)&\multicolumn{3}{c|}{3000 matched faces}&\multicolumn{3}{c|}{3000 dismatched faces}&Accuracy\\
\hline
&Correct&Error&Noface&Correct&Error&Noface&\\
\hline
0.6&2347&649&4&2996&0&4&0.8903\\
0.5&2762&234&4&2996&0&4&0.9597\\
0.4&2925&71&4&2996&0&4&0.9868\\
0.3&2965&31&4&2996&0&4&0.9935\\
0.2&2981&15&4&2996&0&4&0.9961\\
\hline
\end{tabular}
}
\end{center}
\end{table}
As shown in Table \ref{table5}, as the threshold value decreases, the accuracy rate gradually increases. When the threshold is set to 0.2, the accuracy rate can reach a maximum of 99.61\%, and the false rate also gradually decreases. Therefore, a suitable threshold is very significant. At the same time, the result between the same and different face pairs shows that the distinction of intra-class differences depends on the threshold. On the contrary, the difference between the classes is less sensitive to the threshold. As shown in Figure \ref{table46} (b), The accuracy of the method combing our RetinaFace-mnet-faster with ArcFace is increased by 1.4\%.

\subsection{Results}
For the RetinaFace-mnet-faster, we experiment in a single-thread CPU and GPU, and ours performs faster. Comparing with RetinaFace-mnet, the speed of our RetinaFace-mnet-faster is more effective without hurting AP, and the speed of RetinaFace-mnet-faster for 640$\times$480 images on the Tesla P40 is increased by 16.7\% relative to the RetinaFace-mnet, and the speed is increased by 70.2\% on a single-thread CPU thread. The accuracy on the LFW, WIDER FACE, and our datasets are increased by 0.11\%, 0.2\%, and 1.4\%, respectively. According to the appropriate threshold and the strides varying with the resolution, we significantly improve the performance of detection. According to the secondary search method and pre-processing module, we improve the accuracy and speed of face recognition for complex scenes in the video.

\section{Conclusions}
According to the random scales, the multi-scales faces improve the detection performance. The paper proposes RetinaFace-mnet-faster to improve the speed of face detection. For face recognition on videos, this paper analyzes the low quality, tiny faces and blurring because of diverse scenes. Then, this work exploits image enhancement for training and testing to alleviate the problem effectively, respectively. The work further designs the secondary search to improve the accuracy of face recognition.
 \bibliographystyle{splncs03}
 \bibliography{template}
\end{document}